\documentclass[conference]{IEEEtran}
\IEEEoverridecommandlockouts

\usepackage{cite}
\usepackage{amsmath,amssymb,amsfonts}
\usepackage{algorithmic}
\usepackage{graphicx}
\usepackage{textcomp}
\usepackage{xcolor}
\usepackage{url}

\usepackage{xspace}
\usepackage{bm}
\usepackage{graphicx}
\usepackage{subfigure}
\usepackage{subcaption}
\usepackage{float}
\newcommand{\ouralg}{\texttt{WaterAdmin}\xspace}
\newcommand{\ml}{\texttt{ML}\xspace}
\newcommand{\rb}{\texttt{Rule}\xspace}
\newcommand{\llmml}{\texttt{WaterAdmin}\xspace}
\newcommand{\llmmlII}{\texttt{WaterAdmin$_2$}\xspace}
\newcommand{\llmmlIV}{\texttt{WaterAdmin$_4$}\xspace}
\newcommand{\llmmlVI}{\texttt{WaterAdmin$_6$}\xspace}

\usepackage{booktabs}
\usepackage{makecell}

\def\BibTeX{{\rm B\kern-.05em{\sc i\kern-.025em b}\kern-.08em
    T\kern-.1667em\lower.7ex\hbox{E}\kern-.125emX}}
\begin{document}

\title{WaterAdmin: Orchestrating Community Water Distribution Optimization via AI Agents
}

\author{
\IEEEauthorblockN{Jiaqi Wen}
\IEEEauthorblockA{
\textit{University of Houston}\\
}
\and
\IEEEauthorblockN{Pingbo Tang}
\IEEEauthorblockA{
\textit{Carnegie Mellon University}\\
}
\and
\IEEEauthorblockN{Shaolei Ren}
\IEEEauthorblockA{
\textit{University of California, Riverside}\\
}
\and
\IEEEauthorblockN{Jianyi Yang\thanks{* Correspondence to: Jianyi Yang (jyang66@uh.edu).}}
\IEEEauthorblockA{
\textit{University of Houston}\\
}
}

\maketitle

\begin{abstract}
We study the operation of community water systems, where pumps and valves must be scheduled to reliably meet water demands while minimizing energy consumption.
While existing optimization-based methods are effective under well-modeled environments, real-world community scenarios exhibit highly dynamic contexts—such as human activities, weather variations, etc—that significantly affect water demand patterns and operational targets across different zones.
Traditional optimization approaches struggle to aggregate and adapt to such heterogeneous and rapidly evolving contextual information in real time.
While Large Language Model (LLM) agents offer strong capabilities for understanding heterogeneous community context, they are not suitable for directly producing reliable real-time control actions.
To address these challenges, we propose a bi-level AI-agent-based framework, \texttt{WaterAdmin}, which integrates LLM-based community context abstraction at the upper level with optimization-based operational control at the lower level.
This design leverages the complementary strengths of both paradigms to enable adaptive and reliable operation.
We implement \texttt{WaterAdmin} on the hydraulic simulation platform EPANET and demonstrate superior performance in maintaining pressure reliability and reducing energy consumption under highly dynamic community contexts.
\end{abstract}


\section{Introduction}

Community water systems are critical infrastructure for delivering fresh water to meet demand of community consumers \cite{hickey2008water,drinking_water_epa,national1977drinking}. However, water operations are highly energy intensive, requiring substantial electricity for distribution across large-scale hydraulic networks. According to the report of U.S. Environmental Protection Agency (EPA) \cite{energy_efficiency_water}, municipal water systems typically account for approximately 30\% to 40\% of total municipal electricity consumption, and the associated energy costs in the United States amount to billions of dollars annually. As urban infrastructure continues to expand, the demand for electricity and water is steadily increasing. Consequently, achieving reliable and energy-efficient water infrastructure operations has become a critical challenge for the development of sustainable and smart communities.

We consider the community water distribution operations whose objectives include satisfying water demand, maintaining required pressure levels, and minimizing energy costs through the regulation of pumps and valves in the Water Distribution Network (WDN). Existing studies have developed a wide range of optimization-based control approaches \cite{optimizing_pump_energy_efficiency_brentan2024optimizing,pressure_control_leakage_minimization_araujo2006pressure,pump_scheduling_fooladivanda2017energy,energy_efficiency_pump_scheduling_water_supply_luna2019improving,water_pump_optimization_cheh2024water,optimal_operation_oikonomou2018optimal}. 
However, achieving reliable and efficient water operations increasingly requires incorporating diverse and dynamic community context, such as human activities, weather conditions, and electricity grid signals, because these factors strongly influence water demand patterns, operational requirements, and electricity costs. Existing optimization methods typically cannot fully exploit this real-time community information, as it is often unstructured and obtained from various online sources.

Large Language Model (LLM) agents have been utilized to enhance water system operations due to their powerful capabilities in information retrieval, multi-modal information processing, and reasoning \cite{LLM_Water_distribution_network_wang2025leveraging,LLM_assisted_pump_operation_hedaiaty2024ai,LLM_EPANET_water_distribution_goldshtein2025large,LLM_water_resource_management_he2025iwms}. Among them, \cite{LLM_EPANET_water_distribution_goldshtein2025large,LLM_water_resource_management_he2025iwms} utilize LLMs to generate rule-based codes to control WDN components. \cite{LLM_Water_distribution_network_wang2025leveraging,LLM_assisted_pump_operation_hedaiaty2024ai} explore the application of LLM agents to generate system models and to provide scheduling suggestions to human operators. Despite these achievements, there still lacks an AI Agent-based optimizer that can  give real-time control actions for WDN components, such as pump speed and valve positions. One may consider using LLM agents to directly generate control actions, but this introduces several limitations \cite{concerns_LLMs_water_engineering_hosseini2025making,Survey_GenAI_Water_systems_latifigenai}. First, LLM agents may generate physically infeasible actions or actions that violate operational constraints, causing higher safety risks. Moreover, LLM agents often become trapped in local optima when solving optimization problems, potentially resulting in poor performance. Last but not least, the information retrieval and reasoning processes of LLM agents can incur large latency, making it infeasible for real-time water operations that require timely responses.   Therefore, we argue that LLM agents are better suited for high-level planning and strategic guidance, while direct actions for pump or valve control should be output by specialized optimization-based models.

These insights bring the opportunities to combine the complementary strengths of LLM agents and optimization-based approaches, enabling effective integration of community contextual information for reliable and energy-efficient water system operations. To this end, we propose a bi-level operational framework \ouralg  that leverages the reasoning and information integration capabilities of LLM agents together with the resilience of optimization-based methods.
Specifically, we make the following contributions.

\textbullet\quad 
We propose a bi-level framework for water operations. At the higher level, LLM agents abstract real-time community context and human instructions to generate structured operational targets, such as water demand forecasts, location-specific pressure ranges, or nominal water tank levels. To facilitate the LLM reasoning to generate informative operational targets, we perform in-context prompting which incorporate historical instances as in-context prompting and guide the LLM agent to generate target levels based on the in-context information. 

\textbullet\quad 
 At the lower level, we develop an ML-based optimizer that exploits the agent-informed structured operational targets to control pumps and valves in real time.
The model is trained to explicitly utilize these agent-informed signals by simulating operational episodes in EPANET~\cite{EPANET} with true sequences of water usage and operational targets.
To address the challenges posed by non-differentiable simulators such as EPANET, we adopt a training procedure based on zeroth-order optimization to estimate gradient information.
This approach enables the ML-based optimizer to capture WDN dynamics while optimizing operational objectives.

\textbullet\quad 
We conduct numerical evaluations on the EPANET NET3 benchmark~\cite{EPANET} under different component configurations.
The proposed \ouralg is compared against the rule-based control integrated in EPANET and an ML-based optimizer without exploiting LLM agent-based context abstraction.
We further perform ablation studies with different LLM agents and varying windows of LLM-informed targets to assess their impact on performance.
The experimental results demonstrate that integrating LLM agents with ML-based optimizers yields superior performance in both operational reliability and energy efficiency.

\section{Water Distribution Optimization}\label{sec:formulation}
We consider the community water distribution optimization problem whose objective is to satisfy water demands and maintain water pressures with minimized energy costs by regulating the WDN components in real time.    

\textbf{Operational states}. A WDN \cite{public_water_systems_epa,MPC_WDN_wang2017non} can be modeled as directed graph model $\mathcal{G}=\{\mathcal{V},\mathcal{E}\}$ \cite{GNN_state_estimation_water_xing2022graph,Physic-informed_GNN_water_ashraf2024physics}. The set of network nodes $\mathcal{V}$ includes demand nodes, water storage nodes and pump stations. For each node $v\in \mathcal{V}$ at time step $t$, the water pressure is denoted as $h_{v,t}$ (psi). For water storage nodes, the pressure is in proportional to its water level. The water usage at a demand node $v$ at time step $t$ is $d_{v,t}$ ($\mathrm{m}^3/s$).    The network nodes are interconnected by pipelines represented by the directed edges $(u,v)\in\mathcal{E}$. The flow rate from node $u$ to node $v$ at time step $t$ is denoted as $q_{(u,v),t}$ ($\mathrm{m^3/s}$) which is a function of the pressure drop from node $u$ to node $v$. Meters and sensors integrated into water systems enable the real-time measurement of pressure, flow rate, water level, and other essential metrics \cite{pressure_monitoring_iot_water_perez2020design,water_level_sensing_loizou2016water,wireless_sensor_system_waterquality_olatinwo2018energy}. We summarize these state variables using a state vector $s_t$.

\textbf{Control components}.
Pumps and valves are the primary components regulating operational states in water networks \cite{pump_scheduling_fooladivanda2017energy,optimal_scheduling_WDS_singh2019optimal,energy_efficiency_pump_scheduling_water_supply_luna2019improving}. Pumps are deployed at pump stations and of different types including constant-speed and variational-speed pumps. We consider variational speed pumps whose speed can be adjusted to provide varying head for overcoming friction and elevation head loss \cite{VSP_tutterow2004variable}. Valves can be adjusted by actuators to regulate water pressure and flow. The action vector $x_t$ at time step $t$ includes pump activation time (s), pump speed (r/min), and valve positions. Scheduling these components directly impacts energy consumption.

\textbf{Optimization objective}.  The objective of WDN optimization includes the demand satisfaction related loss, energy cost, and operational constraints \cite{MPC_WDN_wang2017non,optimizing_pump_energy_efficiency_brentan2024optimizing,optimal_scheduling_WDS_singh2019optimal}. A demand node or a water tank has a nominal head $h^*_v$ which are expected to be kept during water operation to satisfy the fluctuating demands. The deviation from the the nominal pressure will incur a loss $l_h(x_t)=\frac{1}{|\mathcal{V}|}\sum_{v\in\mathcal{V}}\|h_{v,t}-h^*_v\|$. 
Additionally, the energy cost is expressed as
$l_{e}(x_t) = e_t \cdot E(x_t)$,
where $E(x_t)$ denotes the energy consumption induced by action $x_t$ at time step $t$, and $e_t$ represents the energy price, which may be fixed or time-varying over an episode depending on the energy pricing agreement.
Converting these metrics into the same measure, the multi-objective optimization objective is expressed as $l(x_1,\cdots, x_T)=\gamma_1 l_h(x_t)+\gamma_2 l_{e}(x_t)$.

The operational constraints of WDNs are described below. First, we require the pressure at each concerned demand node and storage tank $v\in\mathcal{V}_D$ satisfies $h_{\min,v}\leq h_{v_,t}\leq h_{\max,v}$ given $h_{\min,v}$ and $h_{\max,v}$ the minimum and maximum pressure limits. The minimum pressure limit $h_{\min,v}$ ensures the satisfaction of water demands while the maximum pressure limit guarantees a safe water pressure. The flow rate of each pipeline should satisfy the physical constraints. For a pipeline $(u,v)\in \mathcal{E}$, it satisfies that $q_{\min,(u,v)}\leq q_{(u,v),t}\leq q_{\max,(u,v)}$ where $q_{\min,(u,v)}$ and $q_{\max,(u,v)}$ are the minimum and maximum flow limits.

\section{Methodology}
In this section, we introduce the design of \ouralg for WDN operations.

\subsection{\ouralg: Framework Overview}
\begin{figure}[!]	
        \centering
        \includegraphics[width={0.45\textwidth}]{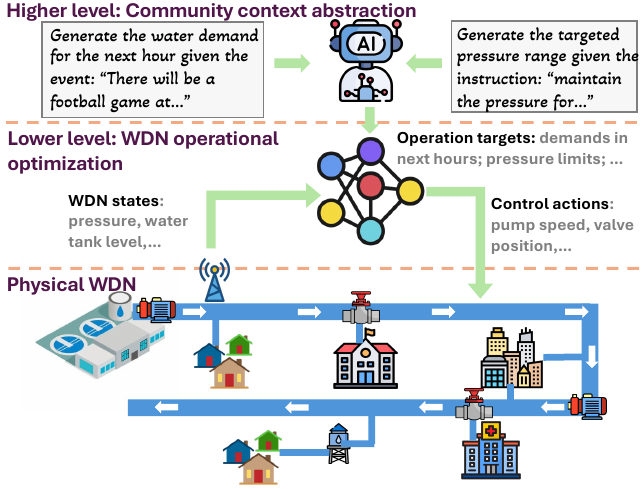}
	\caption{Illustration of \ouralg Architecture.}
	\label{fig:wateradmin}
\end{figure} 
\ouralg is a bi-level framework illustrated in Figure~\ref{fig:wateradmin}. 
At the higher level, LLM agents are used to summarize community context descriptions and generate high-level operational targets.
These context descriptions may include event schedules for major buildings, weather conditions, and other factors that influence water demand patterns across zones. Also, they can include human instructions such as required pressure ranges or flow rates in specific areas to support activities like fire protection~\cite{NFPA_firefighting}.
Based on this information, the LLM agents produce operational targets, including forecasted water demands over the next several time steps, target pressure ranges for critical zones, and desired water tank levels.
These targets are abstracted into a structured format by the LLMs and subsequently processed by the lower-level ML-based optimizer.

At the lower level, an ML model takes as input both the structured operational targets abstracted by the LLMs and the current WDN states.
The WDN states include nodal water pressures, pipeline flow rates, and water tank levels.
Based on this information, the ML model directly outputs control actions for WDN components, including pump speeds and valve positions.
Next, we present the design details for both levels of \ouralg.

\subsection{Context Abstraction via LLM Agents}
LLM agents \cite{LLM_agents_luo2025large,LLM_autonomous_agents_wang2024survey,LLM_multi-agent_guo2024large} have recently emerged as powerful foundation models capable of performing a wide range of tasks, including natural language understanding, information summarization, and decision support across diverse domains.
Leveraging these capabilities, we can directly utilize commercial LLM agent APIs for community context abstraction in the operation of Water Distribution Networks (WDNs), enabling the transformation of heterogeneous and unstructured contextual information into structured operational targets.

Nevertheless, guiding LLMs to generate informative operational targets through prompting remains challenging.
First, LLMs are not well suited for accurately generating numerical quantities, such as continuous water demand values or pressure.
Moreover, it is difficult for LLMs to fully capture complex community scenarios when constrained by prompts of limited length. We solve these challenges by the following methods.

First, to facilitate LLM understanding of numerical quantities in water operations, we discretize them into high-level concepts.
Taking water demand as an example, we partition the continuous demand range into several regions and assign a qualitative level to each region, where Level~1 corresponds to the lowest water usage and higher levels indicate increasing demand.
The partition is not necessarily uniform; in our implementation, we adopt a logarithmic scale to define the regions.
This approach removes the need for LLMs to reason about precise numerical values that may vary across communities, allowing them instead to more effectively capture relative information about water usage levels. 

In addition, to enable effective in-context inference and target generation, we design a set of Chain-of-Thought (CoT) prompts as inputs to the LLM~\cite{CoT_prompting_wei2022chain}.
These prompts incorporate historical context descriptions together with their corresponding operational targets as in-context examples.
For example, in the task of water demand generation, the prompt includes diverse historical instances consisting of event descriptions and their associated water usage levels.
We then append an inference CoT query, such as:
\emph{``Given the event description \dots, generate the water demand level for the next hour at this zone in the format [\dots], and briefly justify the demand by comparing it with historical examples.''}
This prompting strategy guides the LLM to reason over historical cases and more effectively abstract operational targets.
Concrete examples of the proposed prompts are provided in the appendix.

\subsection{ML-based Operational Optimization}
The lower-level optimizer provides direct control actions for WDN components including pumps and valves. The lower-level optimizer must be able to exploit the context abstraction by LLM agents to optimize the operational objectives, achieving reliable and energy efficient water distribution. 

Traditional methods for pump and valve control typically solve large-scale optimization problems based on water distribution formulations~\cite{optimizing_pump_energy_efficiency_brentan2024optimizing,pump_scheduling_fooladivanda2017energy,energy_efficiency_pump_scheduling_water_supply_luna2019improving,water_pump_optimization_cheh2024water,chance_constrained_water_distribution_stuhlmacher2020water}.
However, these optimization-based approaches are largely built upon fixed hydraulic formulations and therefore have limited ability to exploit structured context abstraction for real-time adaptation to dynamic community scenarios.
Moreover, simplified hydraulic dynamic models often deviate significantly from the behavior of real-world water systems.
As water system simulators continue to improve in fidelity~\cite{EPANET,digital_twin_water_system_homaei2024digital}, it becomes increasingly important to incorporate simulation-based knowledge into the optimization process.

To address these challenges, we develop an ML-based optimizer that can leverage structured context abstraction and incorporate simulator knowledge during training.
Specifically, we train an ML-based optimizer $f_{\theta}$, parameterized by $\theta$, by simulating operational episodes—each consisting of $T$ rounds—using EPANET~\cite{EPANET} with the true sequences of water demand and operational target signals, denoted by $\bm{y} = \{y_1, \ldots, y_T\}$.
At each round $t$, the structured operational targets generated by the LLMs are converted into numerical representations $z_t$ that can be processed by the ML model.
Given the WDN state $s_t$ and the target signals $z_t$, the ML model outputs control actions $\bm{x} = \{x_1, \ldots, x_T\}$, where $x_t = f_{\theta}(z_t, s_t)$.

The ML model is trained by minimizing the operational loss
\begin{equation}
    \min_{\theta} \; L(\theta) := l(\bm{x}, \bm{y}) + \lambda^{\top} p(\bm{x}, \bm{y}),
\end{equation}
where $l(\cdot)$ denotes the optimization objective introduced in Section~\ref{sec:formulation}, $p(\cdot)$ is a barrier function that penalizes constraint violations related to pressures or flow rates in Section~\ref{sec:formulation}, and $\lambda$ is the associated Lagrangian weight.
To address the challenges posed by the non-differentiable simulator like EPANET, we train the ML model using zeroth-order optimization.
At each training step, the parameter vector $\theta$ is updated via gradient descent, where the gradient is estimated using a zeroth-order estimator:
\[
\hat g = \mathbb{E}\!\left[\frac{L(\theta + \delta \bm{u}) - L(\theta - \delta \bm{u})}{2\delta} \, \bm{u}\right],
\]
with $\bm{u} \sim \mathcal{N}(0, \bm{I})$, and the expectation taken over the randomness of the direction vector $u$.

\section{Experiment}

In this section, we present the main evaluation results of \ouralg where LLMs are used to predict demand satisfaction targets based on community context.
\subsection{Setups}

\subsubsection{EPANET Setups}: 
In the experiments, we adopt the EPANET benchmark network NET3~\cite{rossman1994epanet} which consists of 97 nodes, including 2 water sources, 3 water tanks, and 92 demand nodes. We select 13 representation nodes which locates at different areas to evaluate the performance. The other details about the EPANET setups are provided in Appendix~\ref{sec:epanet}.

\subsubsection{Objective settings}
 Following the recommendations in the EPA Service Water Pressure Technical Sheet~\cite{EPA_ServiceWaterPressure}, we set the nominal nodal pressure target to 60~psi. A quadratic penalty is applied to quantify deviations of nodal pressure from this nominal value. We further enforce a maximum pressure limit of 100~psi and a minimum pressure limit of 20~psi; pressures exceeding these bounds are treated as violations. The energy consumption at each node and each hour is computed by the EPANET simulator. We evaluate the total  energy consumption per hour across the entire community, without considering time-varying energy prices.

\subsubsection{Demand Datasets}
We adopt the water consumption data from \cite{cheh2024water} to simulate the demand. Details about data pre-processing are provided in Appendix~\ref{sec:datasets}.

\subsubsection{Agent Setups} We provide the setups for LLM agents used for in-context prompting and water demand forecasting.

\textbf{In-context Prompting}.
We employ an LLM agent to construct in-context prompting examples that encode community context and corresponding consumer demand levels. The original demand samples are discretized into five usage levels (0–4), representing increasing water consumption intensities. ChatGPT-4o is used to generate representative pairs, where each example consists of a concise event description and its associated demand level. To this end, the LLM is assigned the role of a domain expert and provided with semantic definitions of the demand levels. It is instructed to generate realistic event narratives by jointly considering usage levels, temporal information, and building types, while capturing characteristic water usage variations. To enhance the diversity of in-context examples, the LLM is allowed to introduce plausible community activities, such as special events, social gatherings, dining promotions, sports events, and seasonal factors.

\textbf{LLM Agents}.
 We employ an LLM agent (ChatGPT-4o by default) to perform water demand forecasting which serves as the demand satisfaction target for lower-level optimization. The LLM is prompted with event-level semantic descriptions and assigned an expert role to infer demand levels over forecasting horizons of 2, 4, or 6 hours based on the provided event descriptions.
The interaction pipeline is implemented using LangChain~\cite{topsakal2023creating}. During the forecasting process, prompt memory is managed via an \texttt{InMemorySaver} checkpointer, which preserves historical dialogue prompts and enables the LLM to maintain contextual continuity across inference steps, even when certain semantic details are not explicitly included in the exemplary prompts.
We evaluate multiple LLM agents, including ChatGPT-4o, DeepSeek-V3, and Gemini-3-flash-preview, and compare their performance in Appendix~\ref{more_results}.

\subsubsection{Baselines}
Our baseline methods include a rule-based optimizer (\rb), a pure ML-based optimizer (\ml), and the proposed method (\llmml). Details about the baselines are provided in Appendix~\ref{sec:baselines}.

\begin{table}[t]
\centering
\caption{Performance comparison of different controllers. }
\label{tab:1}
\begin{tabular}{lcccc}
\toprule
Controller 

& P-MSE 
& \makecell{Max Viol. \\ (\%)} 
& \makecell{Min Viol. \\ (\%)} 
& \makecell{Energy \\ (MWh/hour)} \\
\midrule
\llmmlVI   & \textbf{0.24} & \textbf{7.57}  & \textbf{14.95} & \textbf{0.85}  \\
\llmmlIV  & 0.33 & 10.64 & 19.13 & 1.31  \\
\llmmlII   & 0.33 & 12.99 & 20.46 & 2.03  \\
\ml          & 0.38 & 19.56 & 17.73 & 2.96  \\
\rb         & 5.55 & 93.49 & 3.09  & 30.01 \\

\bottomrule
\end{tabular}
\end{table}
\subsection{Results}  
Table~\ref{tab:1} reports the performance of different approaches on the test demand dataset. Here, P-MSE denotes the mean squared error of pressure deviation from the nominal pressure target and is normalized by the nominal pressure. Max Viol. and Min Viol. represent the violation rates of the upper and lower pressure limits, respectively. The total energy consumption per hour for the entire community is also reported.

The results show that training the ML model to explicitly optimize the operational objective enables the ML-based optimizer to significantly outperform EPANET’s built-in rule-based control in terms of both pressure regulation and energy efficiency. This demonstrates the effectiveness of zeroth-order–trained ML models in capturing the hydraulic dynamics of the simulator and optimizing water distribution objectives.

Moreover, by incorporating LLM-informed targets, \ouralg consistently outperforms the \ml baseline. In particular, compared to \ml, \llmmlVI with a 6-hour prediction window reduces the pressure MSE by 36.8\% and the energy consumption by 71.2\%, highlighting the substantial benefits of integrating LLM-based context abstraction.

Finally, we evaluate \ouralg under different prediction windows to examine the impact of the amount of LLM-provided information. As the prediction window increases, \ouralg achieves improved pressure stabilization and lower energy consumption, indicating that richer LLM-informed context contributes to better water distribution performance.

\section{Conclusion}
In this paper, we study AI-agent–based methods for improving water distribution operations with the objective of satisfying water demands while minimizing energy consumption. We propose a bi-level framework in which the higher level performs LLM-based community context abstraction to generate structured operational targets, and the lower level employs an ML-based optimizer for precise control of WDN components. The proposed framework is evaluated in the EPANET simulation environment, demonstrating the effectiveness of AI agents in stabilizing water pressure and reducing overall energy consumption.

\section{\textbf{Appendix I} \quad Details of Experimental Setups}\label{exp_details}

\subsection{EPANET Setting}\label{sec:epanet}

As shown in Fig.~\ref{fig:net3}, we use the built-in EPANET benchmark network NET3~\cite{rossman1994epanet} as the underlying water distribution system in our experiments, and employ Epyt~\cite{Kyriakou2023} as an interface to connect EPANET with our experimental framework.

\subsubsection{Water net settings}
\begin{figure}[t]
    \centering
    \includegraphics[width=1\linewidth]{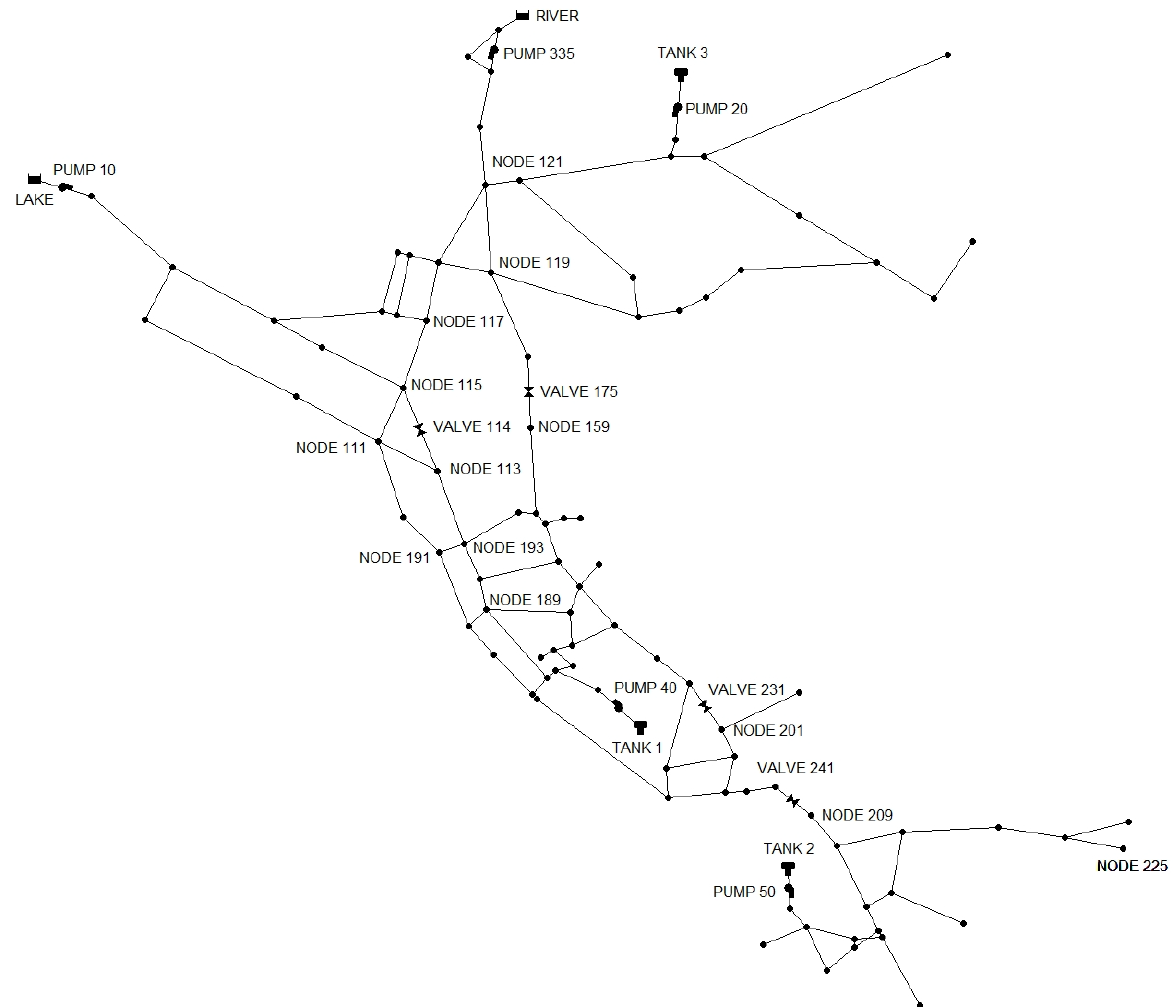}
    \caption{Topology of NET3.}
    \label{fig:net3}
    \end{figure}

All nodal and pipe parameters are kept identical to the original NET3 configuration, except for some pipes that are modified to pumps or valves to enable network control. As illustrated in Fig.~\ref{fig:net3}, the NET3 network consists of 97 nodes, including 2 water sources, 3 water tanks, and 92 demand nodes.

\textbf{Nodes}: We partition the NET3 network into 4 regions with approximately balanced numbers of nodes, where nodes within the same region have similar demand patterns. We select a subset of representative nodes in each region to characterize regional pressure conditions for evaluation purpose. We select nodes with higher connectivity (degree $\geq 4$), nodes located near the geometric center of each region, and junction nodes that serve as critical hubs between neighboring regions. Specifically, we choose 3--4 monitoring nodes per region, resulting in a total set of interest nodes given by Node \texttt{111, 113, 159, 201, 209, 117, 115, 121, 119, 193, 189, 191, 225} in Fig.~\ref{fig:net3}.

\textbf{Pumps}: In NET3, a pump is deployed downstream of each water source and water tank to regulate system inflow and pressure levels. Each pump is characterized by a predefined head--flow performance curve specified by multiple operating points. There are four pumps deployed in our network. The hydraulic characteristics of each pump are specified by a set of discrete head--flow operating points, which are used by EPANET to construct the corresponding pump performance curves. Specifically, pumps located at links \texttt{10}, \texttt{40}, \texttt{50}, and \texttt{20} share an identical characteristic curve defined by three operating points: $(Q,H)=(0,500)$, $(2000,300)$, and $(4000,100)$, where $Q$ denotes the flow rate ($m^3/s$) and $H$ denotes the pump head (m). 
The pump installed at link \texttt{335} is configured with a higher-capacity characteristic curve given by $(Q,H)=(0,500)$, $(8000,138)$, and $(14000,86)$, representing a pump with larger flow capacity and lower head drop rate under high-flow conditions. These heterogeneous pump configurations enable the network to exhibit diverse hydraulic responses under different operating regimes.

For all pumps, the initial relative speed is set to 1.0 (base pump speed), and the allowable speed control range is constrained to $[0.3, 3.0]$, providing sufficient flexibility for adaptive pressure and flow regulation.

\textbf{Valves}: PBV valves provide an effective and direct mechanism for introducing fixed pressure drops to downstream nodes. In the NET3 network, we install PBV valves on four major pipelines including Link~\texttt{114,175,231,241}, which typically serve as hydraulic hub links connecting different regions of the network. By adjusting the PBV valves, controllable pressure drops can be introduced across the selected links, enabling effective regulation of downstream nodal pressure levels. The initial pressure drop of each valve is fixed at 20~psi, and the controllable range is set between 8~psi and 50~psi.

\textbf{Tanks}: The NET3 network includes three storage tanks, and we install pumps on the downstream pipelines of these tanks. However, instead of explicitly regulating tank operations by our methods, we delegate tank-related pump management to EPANET’s built-in hydraulic rule-based control mechanisms.

\subsubsection{Configuration}
    We adopt the pressure-driven demand (PDA) model in EPANET to more realistically capture the coupling between nodal pressure and actual water supply. Specifically, when the nodal pressure is below the minimum threshold $p_{\min}=0$, no water is delivered to the corresponding node, while full demand satisfaction is achieved when the pressure exceeds the required threshold $p_{\mathrm{req}}=60$~psi. For intermediate pressure levels, the demand satisfaction is determined by a nonlinear pressure--demand relationship with an exponent parameter $p_{\mathrm{exp}}=0.5$. 

    The temporal resolution of the simulation is configured with an hourly demand pattern update interval and a one-hour hydraulic time step. The reporting time step is also set to one hour, ensuring that system states and performance metrics are recorded at the same temporal granularity. The total simulation horizon is specified according to the experimental duration setting.
 
\subsection{Datasets}\label{sec:datasets}
    We adopt the water consumption data from~\cite{cheh2024water}, selecting measurements from two academic buildings, one residential area, and one dining facility. The raw data are augmented through scaling to generate sequential demand patterns for each demand node in NET3 (Fig.~\ref{fig:net3}). 
To simulate demand heterogeneity across different areas, we partition the demand nodes in NET3 into four regions and assign nodes within the same region demand profiles corresponding to the same facility type. For example, Region~1 represents a residential area, and all nodes within this region are assigned demand data from the residential dataset. Variations among nodes within the same region are captured through different scaling factors. After data augmentation, the resulting dataset contains 123 days of hourly demand measurements (i.e., $123 \times 24$ time steps) for each node in NET3. We split the dataset chronologically, using the first 70\% episodes of each node for training and the remaining 30\% for testing.

\subsection{Baselines}\label{sec:baselines}
Our baseline methods include a rule-based optimizer (\rb), a pure ML-based optimizer (\ml), and the proposed LLM-enhanced ML-based optimizer (\llmml). 

\textbf{\rb}: \rb applies the EPANET simulator’s built-in rule-based control for pumps and valves, which determines pump activation or shutdown schedules solely based on whether the real-time nodal water demand exceeds the available supply capacity.

\textbf{\ml}: The \ml baseline is implemented using a neural network–based optimization framework with a three-layer fully connected feedforward architecture. The model receives the current nodal pressure states and  the periodic sinusoidal time embedding of $t$ as input and produces continuous control signals for pump speed regulation and valve actuation.

\textbf{\llmml}: The \llmml approach extends the \ml architecture by augmenting the input space with future water demand predictions provided by an LLM agent. Three \llmml variants are evaluated with different forecast horizons, denoted as \llmmlII, \llmmlIV, and \llmmlVI, corresponding to 2-hour, 4-hour, and 6-hour prediction windows, respectively.

\subsection{Additional Details about ML Training}

During training of both \ouralg and the \ml baseline, 
the total number of training epochs is fixed to 10, and the number of samples per iteration in the zeroth-order optimization procedure is also set to 10. The standard deviation of Gaussian noise used in the zeroth-order gradient estimation is configured as $\sigma = 0.05$.

Regarding the learning rate configuration, the \ml baseline uses a learning rate of $2 \times 10^{-5}$, while \llmmlII, \llmmlIV, and \llmmlVI adopt learning rates of $5 \times 10^{-5}$, $2 \times 10^{-4}$, and $5 \times 10^{-4}$, respectively. Since EPANET simulations are executed on the CPU, zeroth-order optimization is performed with a batch size of 1, i.e., each training iteration processes one simulation rollout at a time. Finally, stochastic gradient descent (SGD) is used as the optimizer for all experiments.

\subsection{Prompt Templates}
\begin{figure}
    \centering
    \includegraphics[width=1\linewidth]{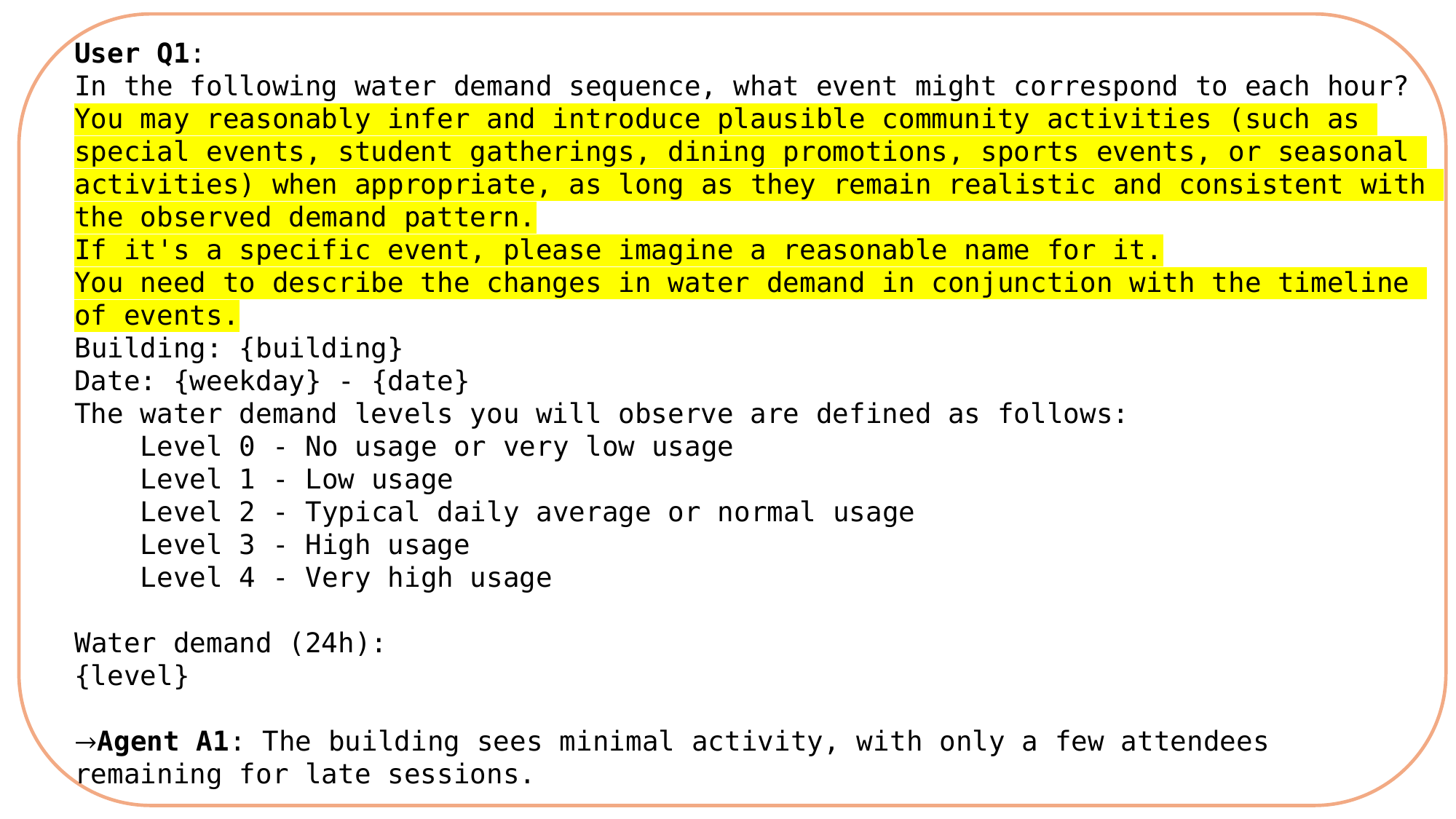}
    \caption{Prompt template for event description generation.}
    \label{fig:generation}
\end{figure}

For event description generation, we use the prompt template illustrated in Fig.~\ref{fig:generation}. In addition to specifying the building type, date information, and water usage level constraints, we explicitly allow the LLM (GPT-4o) to introduce a certain degree of creative inference about human activity patterns in order to enhance the diversity and realism of the generated events. 

\begin{figure}
    \centering
    \includegraphics[width=1\linewidth]{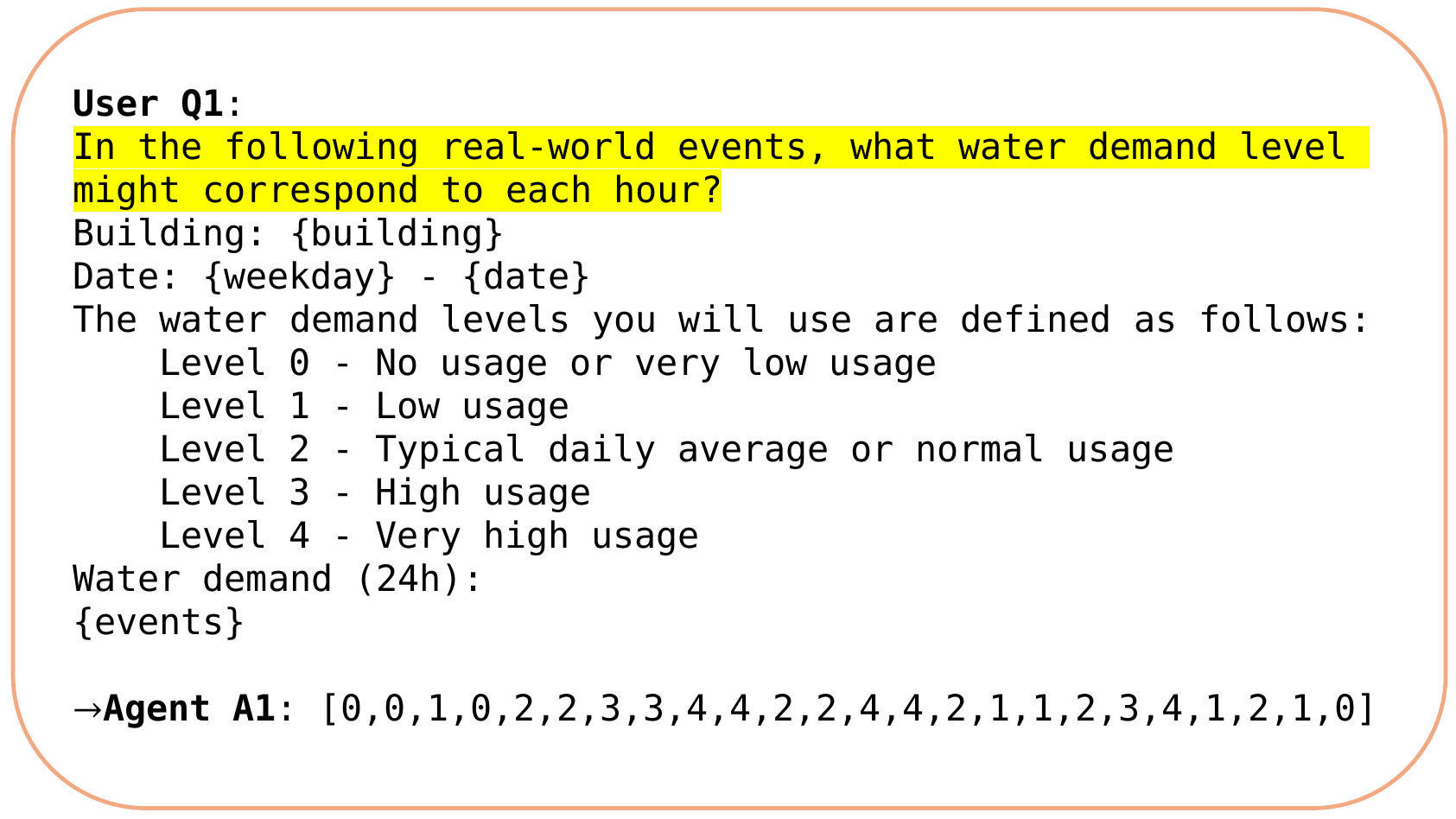}
    \caption{Prompt template for water demand forecasting}
    \label{fig:predict}
\end{figure}

For demand level prediction, to avoid information leakage from the event-generation LLM, we use an LLM API that is different from the one used for event generation. This ensures that no memory or contextual information from the event-generation process is included during demand level prediction and testing. As shown in Fig.~\ref{fig:predict}, we  provide the LLM with the event text, building category, and the same set of classification rules. The enforcement of structured output formats for all LLM responses is automatically handled by the LangChain framework.

\begin{figure*}
    \centering
    \subfigure[A pressure snapshot of Node 113]{
    \includegraphics[width=0.31\textwidth]{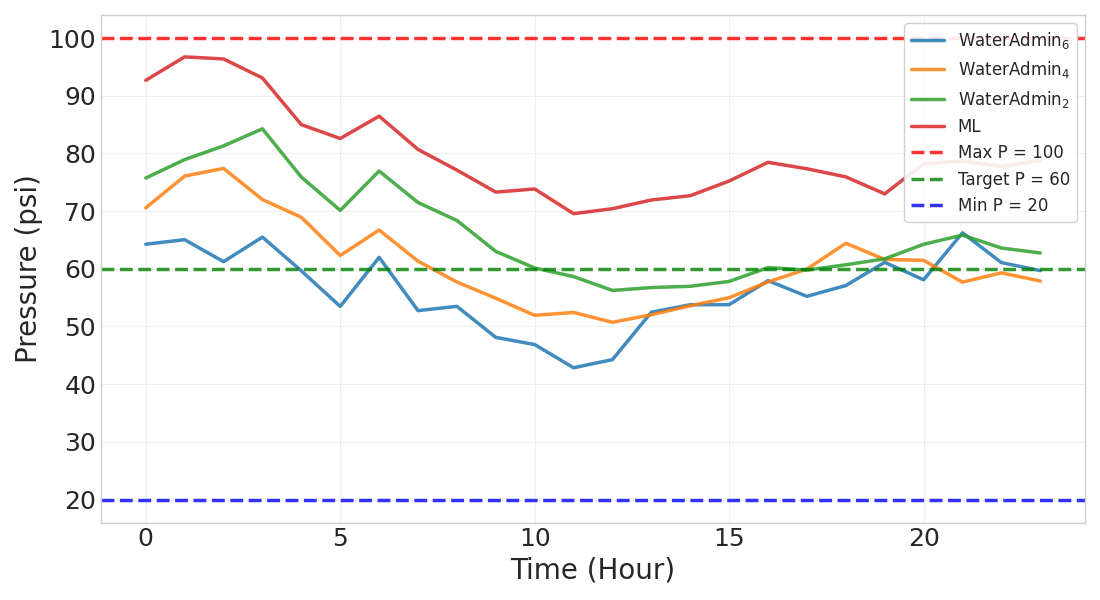}
    \label{fig:node113}
    }
     \subfigure[Pressure distribution of different methods.]{
    \includegraphics[width=0.31\textwidth]{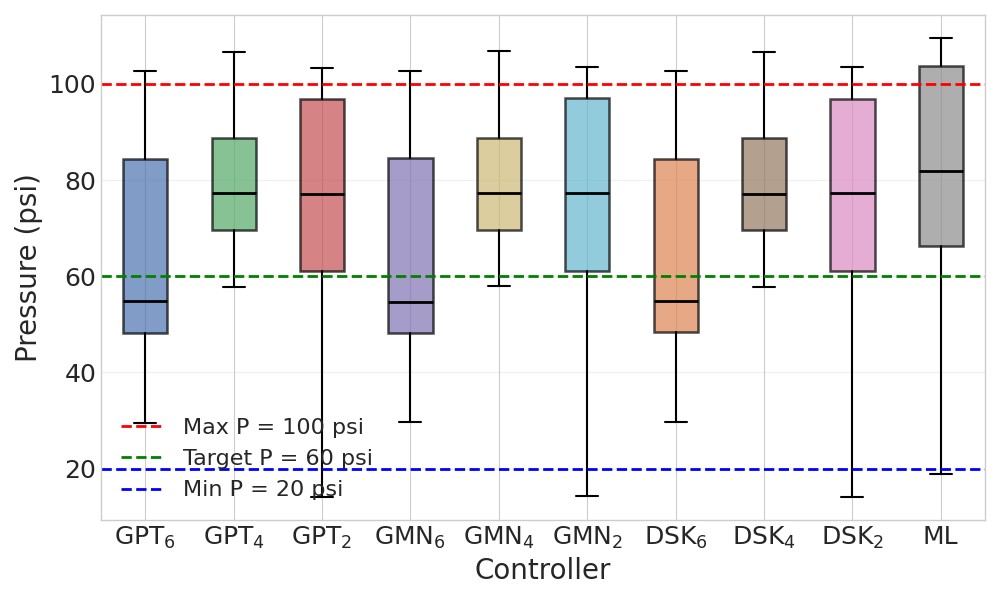}

    \label{fig:pressure}
    }
   \subfigure[Energy consumption distribution of different methods.]{
    \includegraphics[width=0.31\textwidth]{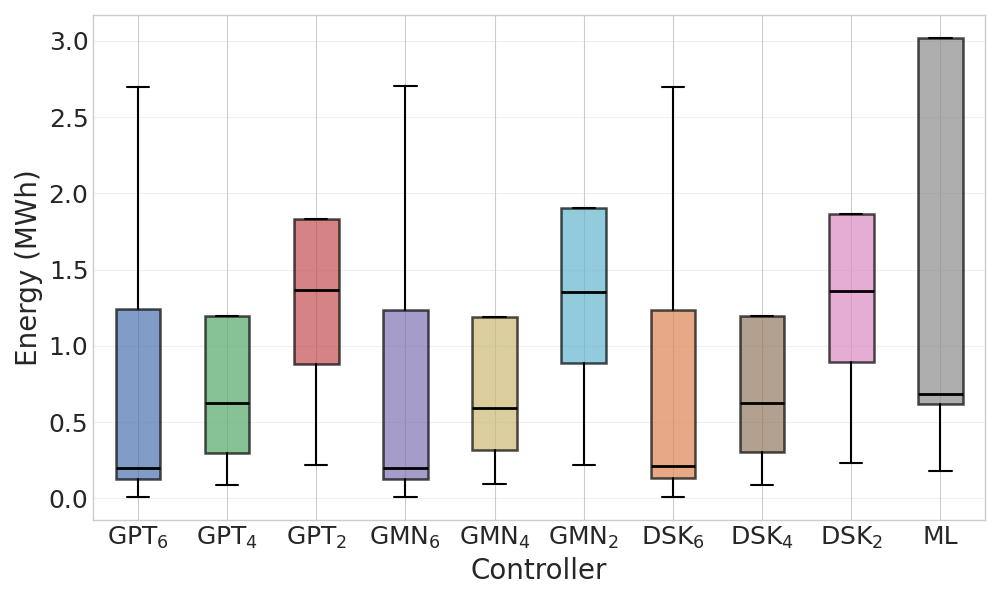}

    \label{fig:energy}
}
\caption{(a) A snapshot of pressure of Node 113 within 24 hours. (b) Pressure distribution of different methods. (c) Energy consumption distribution of different methods. The green dotted line represents the nominal pressure, the blue dotted line represents the low pressure limit, and the red dotted line represents high pressure limit. GPT represent ChatGPT-4o, GMN represents Gemini-3-flash-preview, and DSK represents DeepSeek-V3. The subscriptions represent the prediction windows (2,4, or 6).}
\end{figure*}

\section{\textbf{Appendix II} \quad Additional Results}\label{more_results}

\subsection{Comparison Among LLMs}

As shown in Table~\ref{tab:LLMs}, we compare the performance of \ouralg with different LLM agents with the same prediction window of 6 hours. In this table, GPT represent ChatGPT-4o (about 200 billion parameters), GMN represents Gemini-3-flash-preview (Mixture-of-Experts (MoE) architecture with about 1.2 – 1.7 trillion parameters and 15 – 40 billion active parameters), and DSK represents DeepSeek-V3 (a MoE architecture with 671 billion total parameters and 37B active parameters).   The results suggest that the choice of a LLM backbone with higher capacity cannot contributes a better overall performance for the considered setting. One possible reason is that water demand is discretized into only five levels, which may limit the extent to which the capabilities of larger LLMs can be fully exploited for more precise water demand prediction. Nevertheless, it shows that \ouralg performs very well even with a relatively small LLM ChatGPT-4o, which owes to the in-context prompting technique to select diverse and most representative exemplary prompts and the CoT prompting that guide the LLM agents to perform reasoning. 
\begin{table}[t]
\centering
\caption{Performance comparison of different LLMs. }
\label{tab:LLMs}
\begin{tabular}{lcccc}
\toprule
Controller 

& P-MSE 
& \makecell{Max Viol. \\ (\%)} 
& \makecell{Min Viol. \\ (\%)} 
& \makecell{Energy \\ (MWh/hour)} \\
\midrule
\llmmlVI-GPT  & 0.24 & 7.57 & 14.95 & 0.34 \\
\llmmlVI-GMN   & 0.24 & 7.59 & 14.72 & 0.34 \\
\llmmlVI-DSK & 0.25 & 7.56 & 15.03 & 0.34 \\
\bottomrule
\end{tabular}
\end{table}

\subsection{A Snapshot of Pressure Change}
In Fig.~\ref{fig:node113}, we provide a snapshot of Node~113 in NET3 (Fig.~\ref{fig:net3}) showing the sequence of node pressure regulated by different approaches. We can find that, \ouralg outperforms \ml in maintaining the pressure around 60 psi. This demonstrates the effectiveness of community context abstraction by LLM agents in improving the water distribution operation performance. 

\subsection{Distribution of Pressure and Energy Consumption}
The distributions of the performance metrics including pressure and energy consumption are shown in the box plot of Fig.~\ref{fig:energy}. It provide more details about the performance of different methods among different nodes in various instances.

We can also observe that the impact of different LLM backbones on the performance of \ouralg is not obvious for the considered setting. In contrast, the length of the prediction window plays a more influential role. As the prediction horizon becomes shorter, the control performance degrades, which in turn leads to increased deviation from the nominal pressure and energy consumption.


\begin{thebibliography}{10}

\bibitem{EPANET}
Epanet: Application for modeling drinking water distribution systems.
\newblock \url{https://www.epa.gov/water-research/epanet}, 2024.

\bibitem{public_water_systems_epa}
Information about public water systems.
\newblock
  \url{https://www.epa.gov/dwreginfo/information-about-public-water-systems},
  2024.

\bibitem{drinking_water_epa}
Drinking water distribution system tools and resources.
\newblock
  \url{https://www.epa.gov/dwreginfo/drinking-water-distribution-system-tools-and-resources},
  2025.

\bibitem{energy_efficiency_water}
Energy efficiency for water utilities.
\newblock
  \url{https://www.epa.gov/sustainable-water-infrastructure/energy-efficiency-water-utilities},
  2025.

\bibitem{NFPA_firefighting}
How much water do fire hydrants provide for firefighting.
\newblock
  \url{https://www.nfpa.org/news-blogs-and-articles/blogs/2024/01/12/fire-hydrant-flow},
  2025.

\bibitem{pressure_control_leakage_minimization_araujo2006pressure}
LS~Araujo, H~Ramos, and ST~Coelho.
\newblock Pressure control for leakage minimisation in water distribution
  systems management.
\newblock {\em Water resources management}, 20:133--149, 2006.

\bibitem{Physic-informed_GNN_water_ashraf2024physics}
Inaam Ashraf, Janine Strotherm, Luca Hermes, and Barbara Hammer.
\newblock Physics-informed graph neural networks for water distribution
  systems.
\newblock In {\em Proceedings of the AAAI Conference on Artificial
  Intelligence}, volume~38, pages 21905--21913, 2024.

\bibitem{optimizing_pump_energy_efficiency_brentan2024optimizing}
Bruno Brentan, Filipe Mota, Andrea Menapace, Ariele Zanfei, and Gustavo
  Meirelles.
\newblock Optimizing pump operations in water distribution networks: Balancing
  energy efficiency, water quality and operational constraints.
\newblock {\em Journal of Water Process Engineering}, 63:105374, 2024.

\bibitem{water_pump_optimization_cheh2024water}
Carmen Cheh, Justin Albrethsen, Zhen~Wei Ng, Binbin Chen, Xin Lou, Zaki Masood,
  and David~KY Yau.
\newblock Water pump operation optimization under dynamic market and consumer
  behaviour.
\newblock In {\em Proceedings of the 15th ACM International Conference on
  Future and Sustainable Energy Systems}, pages 335--346, 2024.

\bibitem{cheh2024water}
Carmen Cheh, Justin Albrethsen, Zhen~Wei Ng, Binbin Chen, Xin Lou, Zaki Masood,
  and David~KY Yau.
\newblock Water pump operation optimization under dynamic market and consumer
  behaviour.
\newblock In {\em Proceedings of the 15th ACM International Conference on
  Future and Sustainable Energy Systems}, pages 335--346, 2024.

\bibitem{national1977drinking}
National Research Council (US) Safe Drinking~Water Committee.
\newblock Drinking water and health: Volume 4.
\newblock 1982.

\bibitem{pump_scheduling_fooladivanda2017energy}
Dariush Fooladivanda and Joshua~A Taylor.
\newblock Energy-optimal pump scheduling and water flow.
\newblock {\em IEEE Transactions on Control of Network Systems},
  5(3):1016--1026, 2017.

\bibitem{LLM_EPANET_water_distribution_goldshtein2025large}
Yinon Goldshtein, Gal Perelman, Assaf Schuster, and Avi Ostfeld.
\newblock Large language models for water distribution systems modeling and
  decision-making.
\newblock In {\em World Environmental and Water Resources Congress 2025}, pages
  921--928, 2025.

\bibitem{LLM_multi-agent_guo2024large}
Taicheng Guo, Xiuying Chen, Yaqi Wang, Ruidi Chang, Shichao Pei, Nitesh~V
  Chawla, Olaf Wiest, and Xiangliang Zhang.
\newblock Large language model based multi-agents: A survey of progress and
  challenges.
\newblock {\em arXiv preprint arXiv:2402.01680}, 2024.

\bibitem{LLM_water_resource_management_he2025iwms}
Guo He, Jungang Luo, Feixiong Luo, Xue Yang, Xin Jing, and Huhu Cui.
\newblock Iwms-llm: an intelligent water resources management system based on
  large language models.
\newblock {\em Journal of Hydroinformatics}, 27(11):1685--1702, 2025.

\bibitem{LLM_assisted_pump_operation_hedaiaty2024ai}
Niuosha Hedaiaty~Marzouny and Rebecca Dziedzic.
\newblock Ai-assisted pump operation for energy-efficient water distribution
  systems.
\newblock {\em Engineering Proceedings}, 69(1):3, 2024.

\bibitem{hickey2008water}
Harry~E Hickey.
\newblock Water supply systems and evaluation methods: Volume i: Water supply
  system concepts.
\newblock {\em Federal Emergency Management Agency (FEMA), Washington DC},
  2008.

\bibitem{digital_twin_water_system_homaei2024digital}
MohammadHossein Homaei, Agust{\'\i}n~Javier Di~Bartolo, Mar {\'A}vila, Oscar
  Mogoll{\'o}n-Guti{\'e}rrez, and Andr{\'e}s Caro.
\newblock Digital transformation in the water distribution system based on the
  digital twins concept.
\newblock {\em arXiv preprint arXiv:2412.06694}, 2024.

\bibitem{concerns_LLMs_water_engineering_hosseini2025making}
Seyed~Hossein Hosseini, Babak Zolghadr-Asli, Henrikki Tenkanen, Kaveh Madani,
  Mir~A Matin, Ibrahim Demir, Avi Ostfeld, Vijay~P Singh, and Dragan Savic.
\newblock Making waves: A conceptual framework exploring how large language
  model-based multi-agent systems could reshape water engineering.
\newblock {\em Water Research}, page 125157, 2025.

\bibitem{Kyriakou2023}
Marios~S. Kyriakou, Marios Demetriades, Stelios~G. Vrachimis, Demetrios~G.
  Eliades, and Marios~M. Polycarpou.
\newblock {EPyT: An EPANET-Python Toolkit for Smart Water Network Simulations}.
\newblock {\em Journal of Open Source Software}, 8(92):5947, December 2023.

\bibitem{Survey_GenAI_Water_systems_latifigenai}
Milad Latifi, Ramiz~Beig Zali, and Dragan Savi{\'c}.
\newblock Genai models for urban water systems: Opportunities, challenges, and
  future directions.
\newblock {\em Cambridge Prisms: Water}, pages 1--20.

\bibitem{water_level_sensing_loizou2016water}
Konstantinos Loizou and Eftichios Koutroulis.
\newblock Water level sensing: State of the art review and performance
  evaluation of a low-cost measurement system.
\newblock {\em Measurement}, 89:204--214, 2016.

\bibitem{energy_efficiency_pump_scheduling_water_supply_luna2019improving}
Tiago Luna, Jo{\~a}o Ribau, David Figueiredo, and Rita Alves.
\newblock Improving energy efficiency in water supply systems with pump
  scheduling optimization.
\newblock {\em Journal of cleaner production}, 213:342--356, 2019.

\bibitem{LLM_agents_luo2025large}
Junyu Luo, Weizhi Zhang, Ye~Yuan, Yusheng Zhao, Junwei Yang, Yiyang Gu, Bohan
  Wu, Binqi Chen, Ziyue Qiao, Qingqing Long, et~al.
\newblock Large language model agent: A survey on methodology, applications and
  challenges.
\newblock {\em arXiv preprint arXiv:2503.21460}, 2025.

\bibitem{optimal_operation_oikonomou2018optimal}
Konstantinos Oikonomou and Masood Parvania.
\newblock Optimal coordination of water distribution energy flexibility with
  power systems operation.
\newblock {\em IEEE Transactions on Smart Grid}, 10(1):1101--1110, 2018.

\bibitem{wireless_sensor_system_waterquality_olatinwo2018energy}
Segun~O Olatinwo and Trudi-H Joubert.
\newblock Energy efficient solutions in wireless sensor systems for water
  quality monitoring: A review.
\newblock {\em IEEE Sensors Journal}, 19(5):1596--1625, 2018.

\bibitem{pressure_monitoring_iot_water_perez2020design}
Jos{\'e} P{\'e}rez-Padillo, Jorge Garc{\'\i}a~Morillo, Jos{\'e} Ramirez-Faz,
  Manuel Torres~Rold{\'a}n, and Pilar Montesinos.
\newblock Design and implementation of a pressure monitoring system based on
  iot for water supply networks.
\newblock {\em Sensors}, 20(15):4247, 2020.

\bibitem{rossman1994epanet}
Lewis~A Rossman et~al.
\newblock Epanet users manual.
\newblock 1994.

\bibitem{optimal_scheduling_WDS_singh2019optimal}
Manish~K Singh and Vassilis Kekatos.
\newblock Optimal scheduling of water distribution systems.
\newblock {\em IEEE Transactions on Control of Network Systems}, 7(2):711--723,
  2019.

\bibitem{chance_constrained_water_distribution_stuhlmacher2020water}
Anna Stuhlmacher and Johanna~L Mathieu.
\newblock Water distribution networks as flexible loads: A chance-constrained
  programming approach.
\newblock {\em Electric Power Systems Research}, 188:106570, 2020.

\bibitem{topsakal2023creating}
Oguzhan Topsakal and Tahir~Cetin Akinci.
\newblock Creating large language model applications utilizing langchain: A
  primer on developing llm apps fast.
\newblock In {\em International conference on applied engineering and natural
  sciences}, volume~1, pages 1050--1056, 2023.

\bibitem{VSP_tutterow2004variable}
Vestal Tutterow and Aimee~T McKane.
\newblock Variable speed pumping: A guide to successful applications.
\newblock 2004.

\bibitem{EPA_ServiceWaterPressure}
{United States Environmental Protection Agency}.
\newblock {\em Service Water Pressure Technical Sheet}.
\newblock EPA, 2023.

\bibitem{LLM_Water_distribution_network_wang2025leveraging}
Jian Wang, Guangtao Fu, and Dragan Savic.
\newblock Leveraging large language models for automating water distribution
  network optimization.
\newblock {\em Water Research}, page 124536, 2025.

\bibitem{LLM_autonomous_agents_wang2024survey}
Lei Wang, Chen Ma, Xueyang Feng, Zeyu Zhang, Hao Yang, Jingsen Zhang, Zhiyuan
  Chen, Jiakai Tang, Xu~Chen, Yankai Lin, et~al.
\newblock A survey on large language model based autonomous agents.
\newblock {\em Frontiers of Computer Science}, 18(6):186345, 2024.

\bibitem{MPC_WDN_wang2017non}
Ye~Wang, Vicen{\c{c}} Puig, and Gabriela Cembrano.
\newblock Non-linear economic model predictive control of water distribution
  networks.
\newblock {\em Journal of Process Control}, 56:23--34, 2017.

\bibitem{CoT_prompting_wei2022chain}
Jason Wei, Xuezhi Wang, Dale Schuurmans, Maarten Bosma, Fei Xia, Ed~Chi, Quoc~V
  Le, Denny Zhou, et~al.
\newblock Chain-of-thought prompting elicits reasoning in large language
  models.
\newblock {\em Advances in neural information processing systems},
  35:24824--24837, 2022.

\bibitem{GNN_state_estimation_water_xing2022graph}
Lu~Xing and Lina Sela.
\newblock Graph neural networks for state estimation in water distribution
  systems: Application of supervised and semisupervised learning.
\newblock {\em Journal of Water Resources Planning and Management},
  148(5):04022018, 2022.

\end{thebibliography}
\end{document}